\pdfoutput=1

\documentclass[11pt]{article}

\usepackage{amsmath,amsfonts,bm}

\def\eqref#1{equation~\ref{#1}}

\def\1{\bm{1}}

\DeclareMathAlphabet{\mathsfit}{\encodingdefault}{\sfdefault}{m}{sl}
\SetMathAlphabet{\mathsfit}{bold}{\encodingdefault}{\sfdefault}{bx}{n}

\DeclareMathOperator*{\argmax}{arg\,max}
\DeclareMathOperator*{\argmin}{arg\,min}

\usepackage[preprint]{acl}
\usepackage{times}
\usepackage{latexsym}
\usepackage{booktabs}
\usepackage{arydshln}

\usepackage{times}
\usepackage{float}
\usepackage{latexsym}
\usepackage{amsthm}
\usepackage{algorithm,algorithmic}
\usepackage{amsmath}
\usepackage{amssymb}
\usepackage{mathtools}

\usepackage{enumitem}
\setlist[itemize]{noitemsep, topsep=0pt, leftmargin=11pt}
\setlist[enumerate]{noitemsep, topsep=0pt, leftmargin=11pt}
\usepackage{wrapfig}
\usepackage{xcolor}
\usepackage{subcaption}
\usepackage{multirow}

\usepackage[T1]{fontenc}

\usepackage[utf8]{inputenc}

\usepackage{microtype}

\usepackage{inconsolata}

\usepackage{graphicx}

\title{Improving Task Diversity in Label Efficient Supervised Finetuning of LLMs}

\author{
  Abhinav Arabelly\thanks{Equal contribution.} \quad
  Jagrut Nemade\footnotemark[1] \quad
  Robert D. Nowak \quad
  Jifan Zhang \\
  University of Wisconsin--Madison \\
}

\begin{document}
\maketitle
\begin{abstract}
Large Language Models (LLMs) have demonstrated remarkable capabilities across diverse domains, but developing high-performing models for specialized applications often requires substantial human annotation — a process that is time-consuming, labor-intensive, and expensive. In this paper, we address the label-efficient learning problem for supervised finetuning (SFT) by leveraging task-diversity as a fundamental principle for effective data selection.
This is markedly different from existing methods based on the prompt-diversity.  Our approach is based on two key observations: 1) task labels for different prompts are often readily available; 2) pre-trained models have significantly varying levels of confidence across tasks.  We combine these facts to devise a simple yet effective sampling strategy: we select examples across tasks using an inverse confidence weighting strategy. This produces models comparable to or better than those trained with more complex sampling procedures, while being significantly easier to implement and less computationally intensive. Notably, our experimental results demonstrate that this method can achieve better accuracy than training on the complete dataset (a 4\% increase in MMLU score). Across various annotation budgets and two instruction finetuning datasets, our algorithm performs at or above the level of the best existing methods, while reducing annotation costs by up to 80\%.
\end{abstract}

\section{Introduction}

Large Language Models have demonstrated remarkable capabilities across a diverse range of tasks and domains. However, for challenging tasks where LLMs still struggle, finetuning often requires a large number of human annotations and demonstrations to guide models toward correct behavior~\citep{qin2024unleashing,zhou2025bridging,chen2025sft}. This annotation process is typically time-consuming, labor-intensive, and expensive, creating a significant barrier to developing high-performing models for specialized applications.

In this work, we address the label-efficient learning problem, where ground-truth responses are initially unknown. The goal is to develop methods that reduce annotation requirements while improving model performance. Our approach leverages task-level information as a fundamental organizing principle for effective data selection in instruction tuning.

The label-efficient SFT~\citep{bhatt2024experimental} and data selection literature~\citep{yang2024smalltolarge,wang2024diversity} have introduced various diversity-based methods based on facility location, clustering, optimal design, and determinantal point process methods to ensure comprehensive data coverage. Despite their improved performances, these approaches face several practical challenges: (1) they typically rely on pre-computed embeddings that inadequately capture task-specific semantics; (2) their computational complexity scales poorly with dataset size, limiting applicability to large-scale problems; and (3) achieving optimal performance often requires careful hyperparameter tuning, adding complexity to implementation.

Our study introduces a novel approach that combines task-diversity principles with uncertainty quantification. Unlike existing methods that focus primarily on prompt-diversity in the embedding space, we leverage the inherent task categorizations \citep{kung2023active,ivison2022dataefficient,wang2022supernaturalinstructions,wei2022finetuned}
widely available in modern LLM development datasets. Prompt here refers to individual input examples to an LLM. We define tasks according to their original data curation sources, such as various domains (e.g., medical knowledge, mathematics) and desired skills (e.g., summarization, translation). For instance, the FLAN dataset \citep{longpre2023flan} combines instruction tuning data from over a hundred sources across 1,691 tasks, while the Dolly dataset~\citep{conover2023free} covers eight distinct tasks, such as brainstorming, closed QA, information extraction and others.

Our approach is straightforward yet effective: we allocate a minimum amount of allocation budget across all tasks to ensure diversity, while allocating more budget to uncertain tasks, using an inverse confidence weighting strategy. The confidence is derived from the base model's confidence in answering the questions. This method achieves high diversity in data selection while prioritizing tasks where the model exhibits greater uncertainty, producing models that outperform those trained with existing sampling procedures, while being significantly more accessible and easier to implement.

Experimental results demonstrate that by prioritizing the labeling of a strategic subset of examples, our method achieves better accuracy than training on the complete dataset (a 4\% increase in MMLU score). Across various annotation budgets and two instruction finetuning datasets, our algorithm consistently performs at or above the level of the best existing methods, while reducing annotation costs by up to 80\%. These findings highlight the effectiveness of combining task diversity with uncertainty quantification as a guiding principle for data selection in instruction tuning and provide a practical approach for optimizing the annotation process without sacrificing model performance.

\section{Related Work}

\subsection{LLM Supervised Finetuning}
Supervised finetuning (SFT) has become a pivotal technique for aligning Large Language Models (LLMs) with human preferences and specific tasks \citep{ouyang2022training, wei2022finetuned, touvron2023llama}. This approach enhances pre-trained language models by training them on carefully curated prompt-response pairs, enabling more accurate instruction following and appropriate response generation. However, traditional SFT methods typically demand extensive human-annotated examples, creating resource-intensive processes that potentially limit scalability. This constraint has spurred research into more efficient finetuning approaches that can achieve comparable performance with significantly fewer labeled examples.

\subsection{Label-Efficient Learning for LLMs}
The pursuit of label-efficient learning for LLMs has garnered substantial attention as researchers strive to reduce annotation burdens while maintaining model performance. Various methodologies have emerged, including active learning and experimental design (i.e., one-batch active learning).
In this paper, we focus on the one-batch active learning problem, similar to \citet{bhatt2024experimental}, where the algorithm selects a single set of informative examples for annotation. This approach offers distinct advantages: it reduces logistical complexities compared to iterative active learning and provides computational benefits by eliminating expensive retraining cycles for LLMs.

\noindent\textbf{Distinction from Data Selection.}
It is essential to distinguish label-efficient learning from data selection. While data selection operates with prior knowledge of ground truth responses, label-efficient learning functions without access to these annotations. Numerous data selection methods \citep{chen2023alpagasus, bukharin2023data, du2023mods, yang2024smalltolarge, wang2024diversity} incorporate ground truth responses as integral components of their selection procedures, rendering them unsuitable for scenarios where annotations have not yet been collected. Our approach addresses this limitation by focusing on task characteristics that can be evaluated prior to annotation.

\noindent\textbf{Unifying Principles.}
Although data selection methods cannot be directly applied to label-efficient settings, certain algorithmic principles are shared between these problem domains. Existing methods in active learning aim to either enhance diversity among selected examples or reduce model uncertainty by prioritizing the most uncertain or incorrectly predicted examples~\citep{lewis1995sequential,tong2001support,settles2009active,kremer2014active,ash2019deep,ash2021gone,citovsky2021batch,mohamadi2022making,zhang2022galaxy,nuggehalli2023direct,zhanglabelbench,xie2024deep}. Below, we provide a concise overview of key algorithms in this space for LLMs.

\begin{figure*}[ht]
    \centering
    \begin{subfigure}[b]{0.32\textwidth}
        \centering
        \includegraphics[width=\linewidth]{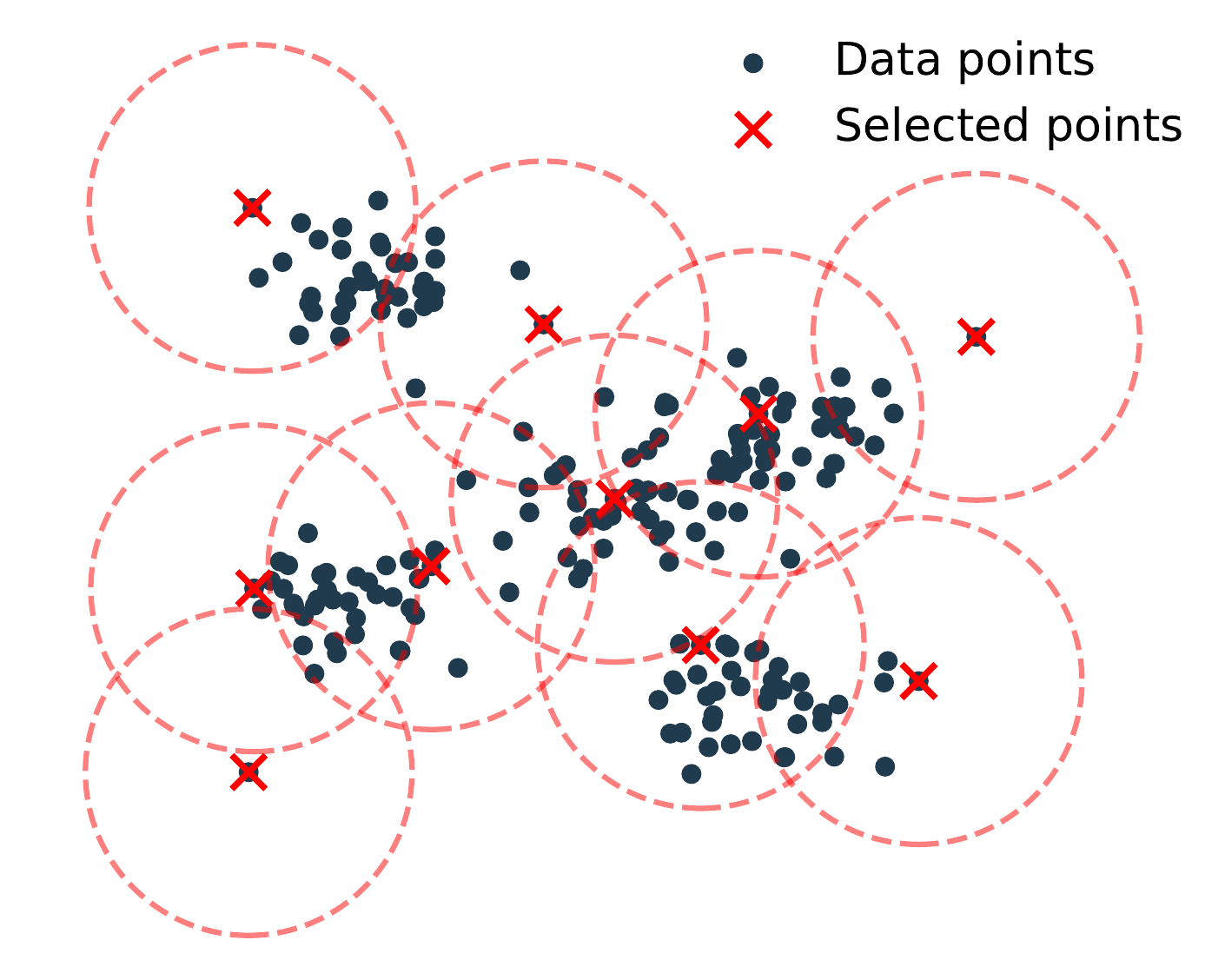}
        \caption{K-Center selection}
        \label{fig:kcenter}
    \end{subfigure}
    \hfill
    \begin{subfigure}[b]{0.32\textwidth}
        \centering
        \includegraphics[width=\linewidth]{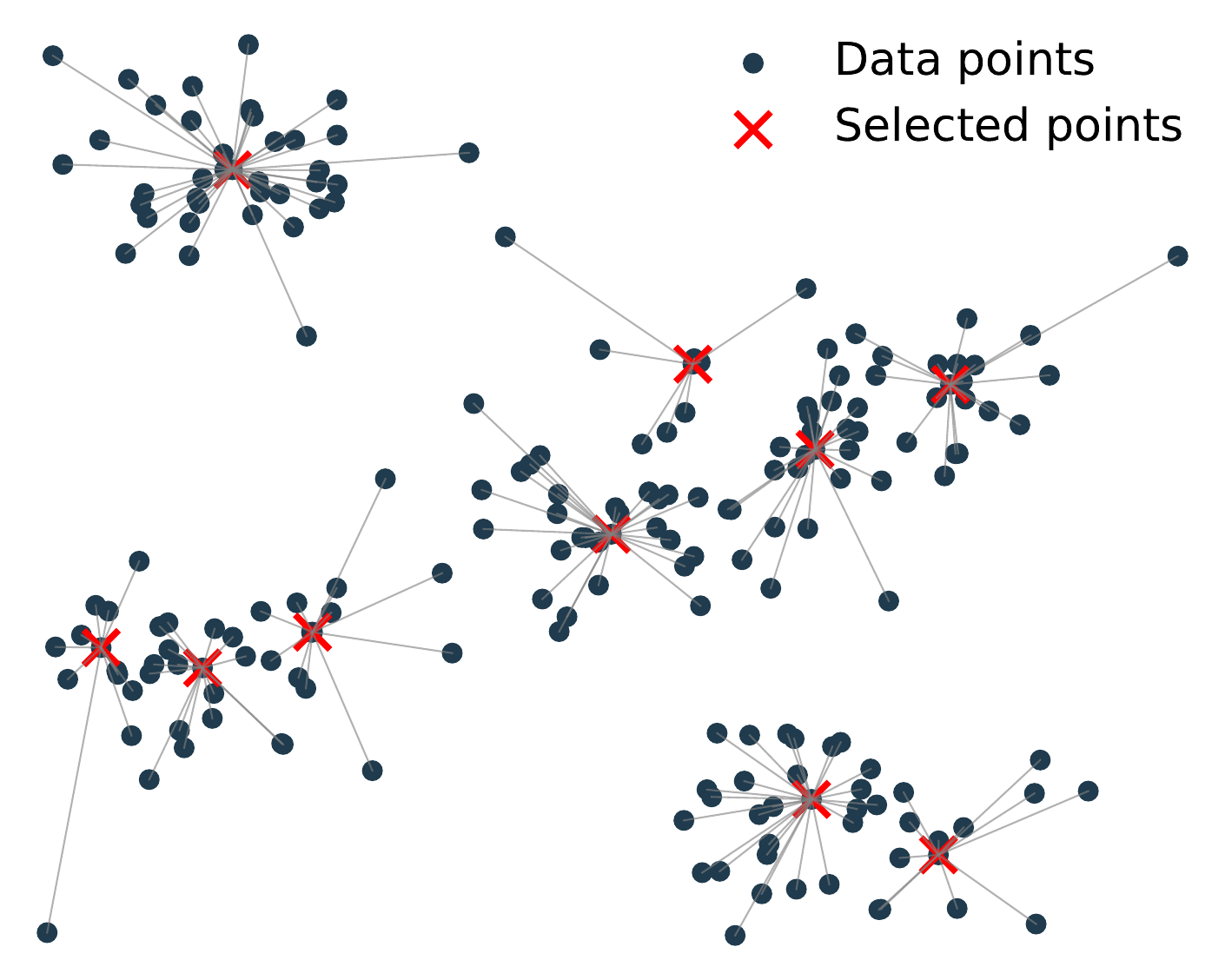}
        \caption{Facility Location selection}
        \label{fig:facility_location}
    \end{subfigure}
    \hfill
    \begin{subfigure}[b]{0.32\textwidth}
        \centering
        \includegraphics[width=\linewidth]{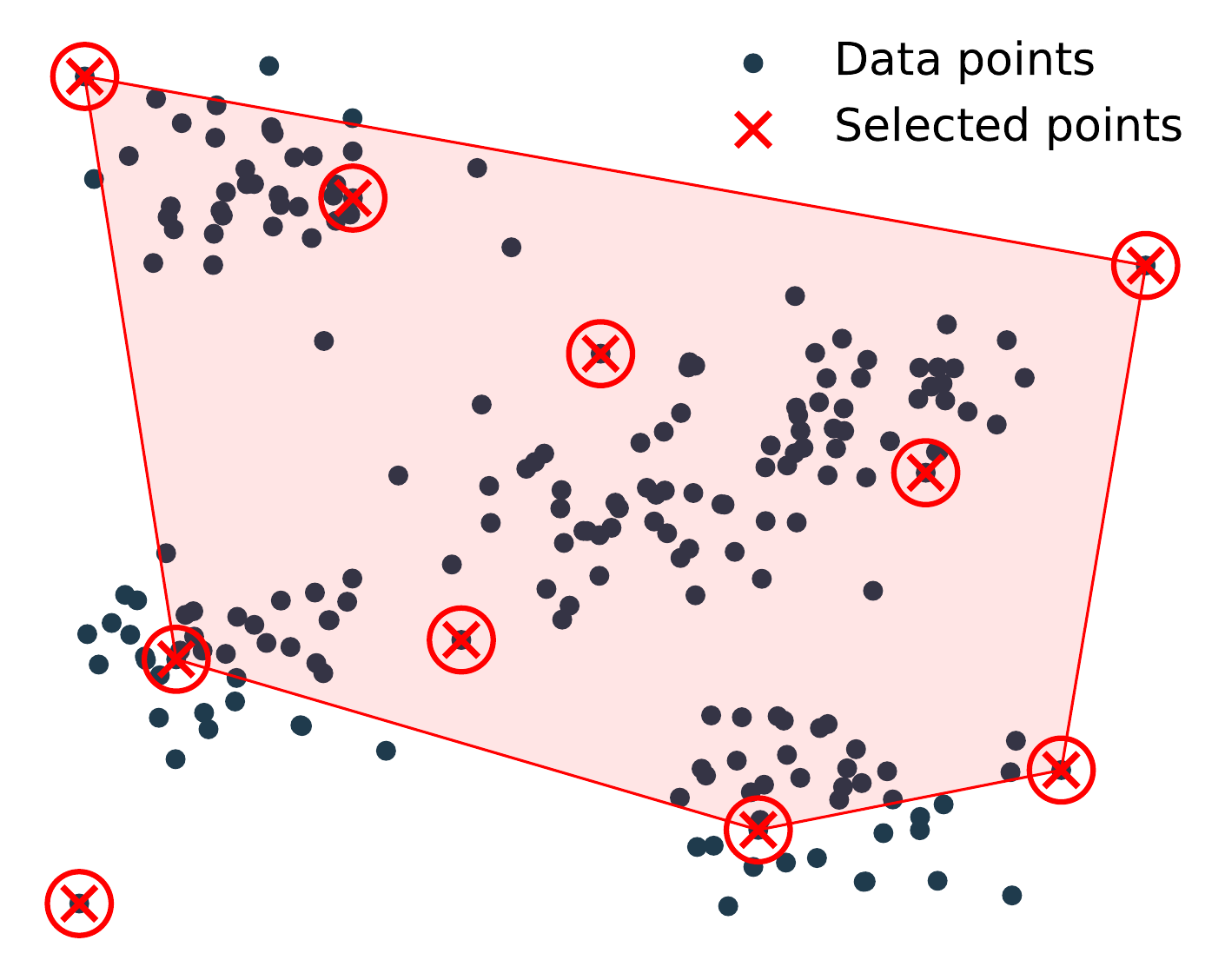}
        \caption{DPP selection}
        \label{fig:dpp_selection}
    \end{subfigure}
    \caption{Comparison of three diversity-based data selection baseline algorithms.
    In contrast to task diversity in Figure~\ref{fig:allocation_strategies}, these methods primarily rely on prompt diversity in the embedding space. All of these methods aim to cover the space of examples based on different criterions.}
    
    \label{fig:selection_methods_grid}
\end{figure*}

\subsubsection{Diversity-Based Methods}\label{ssec:div_review}
Diversity-based data selection aims to construct representative subsets that ensure broad coverage of the input space by prioritizing diversity among the prompts selected for annotation.
Below, we review three prominent diversity-based selection strategies: K-Center, Facility Location, and Determinantal Point Processes (DPPs). We use $Z = z_1, ..., z_m$ to denote the set of data examples in some embedding space, and the goal is to select the subset $S \subset Z$ to ensure good coverage of examples.

\textbf{K-Center} \citep{sener2018coreset} selects $k$ data points to serve as centers of equal-radius balls in the representation space. The objective is to cover all data points using the smallest possible radius, ensuring that each example lies within the ball of at least one selected center (see Figure~\ref{fig:kcenter}). This approach encourages the selection of a subset that minimizes the maximum distance between any data point and its closest selected center. Formally, the objective function is:
\begin{align*}    
S = \arg\min_{\substack{S' \subset X \\ |S'| = k}} \; \max_{z_i \in Z} \; \min_{z_j \in S'} \; \|z_i - z_j\|
\end{align*}
As illustrated in Figure~\ref{fig:kcenter}, this coreset sampling method is vulnerable to outliers, as remote singular examples are often selected.

\textbf{Facility Location (FL)} \citep{mirchandani1990discrete, wei2015submodularity, mirzasoleiman2020coresets, bilmes2022submodularity, bukharin2023data, bhatt2024experimental} addresses the outlier issue by considering the average distance of examples to their corresponding centers rather than the worst case distance. This strategy utilizes a similarity kernel measure as a proxy for distance metrics, denoted by $\mathcal{K}(\cdot, \cdot)$, which quantifies the similarity between features $z_i$ and $z_j$. The facility location objective is formulated as:
\begin{align*}
S = \argmax_{\substack{S' \subset X \\ |S'| = k}}\; \sum_{z_i \in X} \;\max_{z_j \in S'} \;\mathcal{K}(z_i, z_j)
\end{align*}
In this formulation, each data point $z_i \in X$ (considered a "client") is assigned to its most similar center $z_j \in S'$ (the "facility"), and the objective maximizes the cumulative similarity across all such assignments. The kernel function can be an $l_2$ distance as in k-center selection, or alternatively an RBF kernel with hyperparameter $\gamma$ \citep{bhatt2024experimental}.

\textbf{Determinantal Point Processes (DPP)} provide a probabilistic framework for subset selection that maximizes data coverage \citep{kulesza2012determinantal, wang2024diversity}. A DPP is inherently a stochastic process where the probability of selecting subset $S$ is proportional to:
\begin{align*}
\det(K(S)) := \det([z]_{z\in S}^\top[z]_{z\in S})
\end{align*}
where $K(S)_{i,j} = \langle z_i, z_j \rangle$ for the $i$-th and $j$-th elements in $S$. This determinant quantifies the volume of the parallelotope formed by the embeddings in $S$ when the selection budget $k$ is no larger than the feature dimensionality $d$. Beyond inner products, the kernel map can be replaced with other kernels $K(S)_{i,j} = \mathcal{K}(z_i, z_j)$ similar to FL. In \citet{wang2024diversity}, the authors employ a deterministic version of DPP, equivalent to D-optimal design: $\argmax_S \log\det(\mathcal{K}(S))$. As shown in Figure~\ref{fig:dpp_selection}, this method tends to select examples along the boundary of the prompt distribution. For data selection, \citet{wang2024diversity} also incorporates quality scores based on ground-truth answers. Since we lack access to such information in label-efficient SFT, we instead integrate confidence scores as proposed by \citet{bhatt2024experimental} throughout our experiments.

\subsubsection{Uncertainty-Based Methods}
Uncertainty-based selection strategies identify examples where the model exhibits low confidence, utilizing metrics such as entropy, token probability, and margin-based measures to quantify prediction uncertainty \citep{settles2009active, lewis1995sequential}. Recent research has explored more sophisticated approaches specifically tailored for large language models, advancing beyond simple uncertainty estimates. \citet{bhatt2024experimental} extended entropy-, confidence-, and margin-based scores to SFT of LLMs. Similar to us, \citet{kung2023active} also utilizes the task definitions from existing datasets, and developed methods that prioritize labeling tasks with the highest uncertainties. However, their algorithm would spend all of their budgets on the most difficulty tasks, thus reducing the data coverage and diversity in selected prompts.
Our work draws inspiration from these techniques while introducing a simpler and more practical approach that focuses on task diversity and model confidence. This provides an effective method for optimizing annotation resources without requiring complex embeddings or algorithms.

\section{Problem Formulation}

We adopt the label-efficient SFT framework introduced by \citet{bhatt2024experimental}. In this framework, the learner is provided with an initial set of $N$ prompts $X = \{x_1, x_2, \dots, x_N\}$, where each prompt $x_i$ consists of $l$ tokens, i.e., $x_i = \{x_{i,1}, \dots, x_{i,l} \}$. Let $g$ represent the pretrained language model. Given a limited annotation budget $k < N$, our goal is to develop effective selection strategies that identify a subset $S \subset X$ of the most informative prompts, such that $|S| = k$.

After selecting subset $S$, high-quality responses are collected for each prompt $x \in S$ from annotators (such as human experts or advanced LLMs). These annotated prompt-response pairs are then used to finetune model $g$. Let $g'$ denote the resulting model after SFT on these annotated pairs. Our objective is to maximize the performance improvement of $g'$ while minimizing the required annotation budget.

Throughout this paper, we use feature embeddings extracted from prompts using the base model. Let $f: X \rightarrow \mathbb{R}^d$ denote the feature mapping between prompts  to embedding, all of the diversity based methods in Section~\ref{ssec:div_review} can be viewed as using $z_i := f(x_i)$. Concretely, we extract these feature embeddings from the penultimate layer's hidden state of the pretrained model $g$ during inference on the prompt.

\begin{figure*}[t]
    \centering

    \begin{subfigure}[t]{0.48\textwidth}
        \centering
        \includegraphics[width=\textwidth]{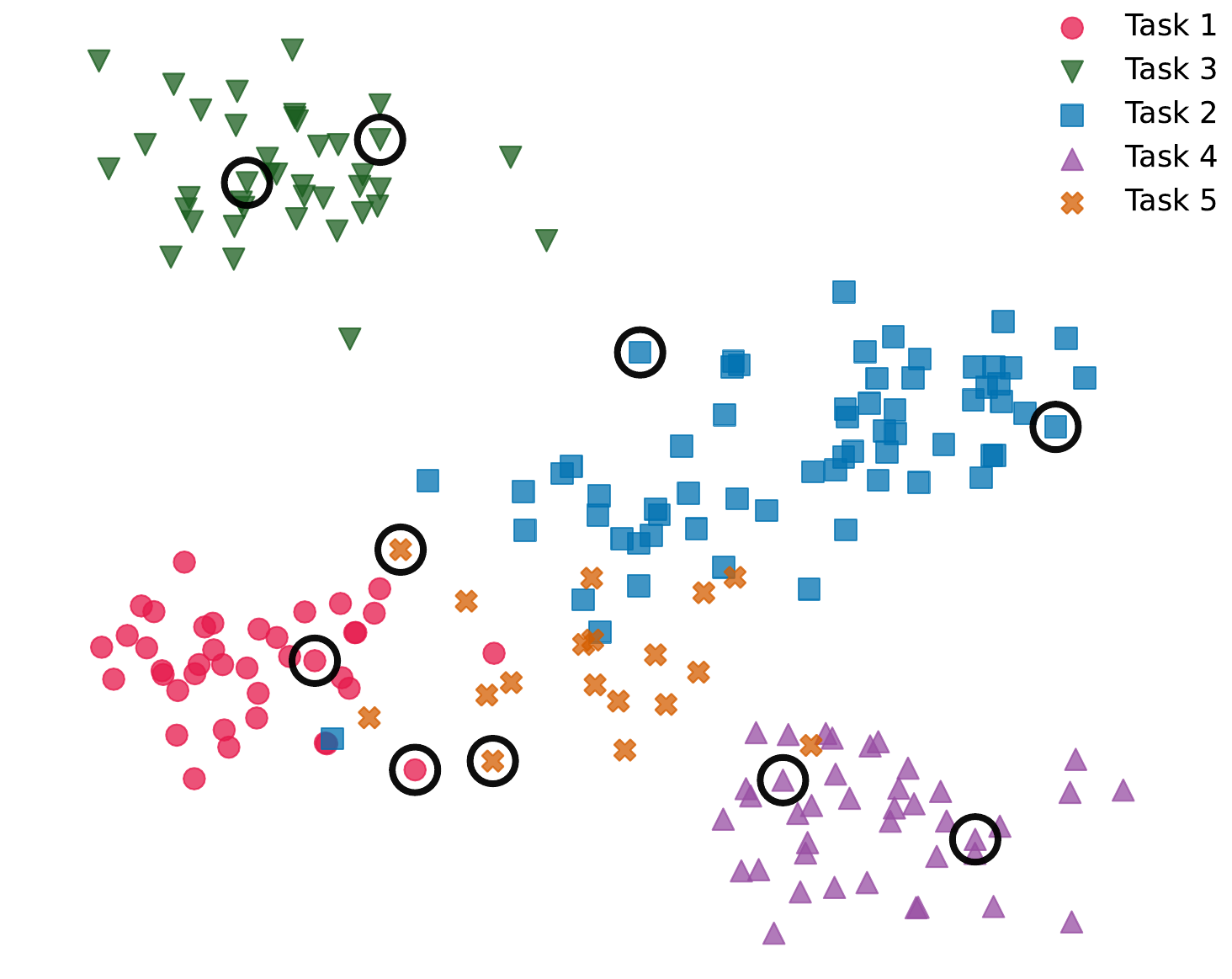}
    \end{subfigure}%
    \hfill
    \begin{subfigure}[t]{0.48\textwidth}
        \centering
        \includegraphics[width=\textwidth]{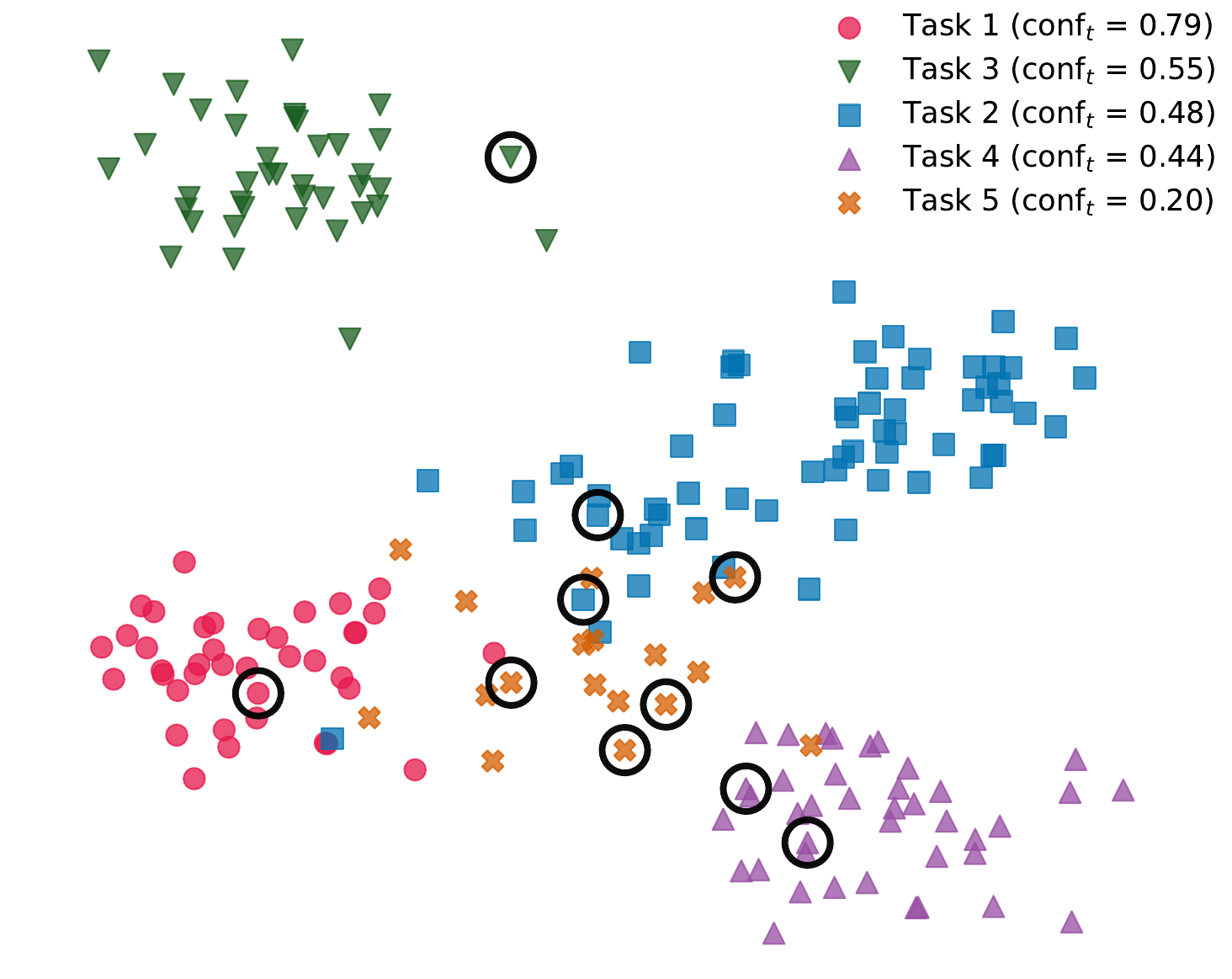}
    \end{subfigure}

    \caption{(a) Task diversity:  Fixed number of examples are selected in a round-robin manner across tasks, without considering the model's confidence of the task.
    (b) Weighted Task Diversity: Examples are allocated across tasks in proportion to the inverse of the average model confidence on each task (denoted by $\text{conf}_t$ for task $t$). Tasks for which the model exhibits lower confidence receive more samples. Despite Task 5 being less represented in the original distribution, we collect more samples because the pre-trained model is generally less confident about the task. When comparing to Figure~\ref{fig:selection_methods_grid}, our method leverages the additional task information for selection.}
    \label{fig:allocation_strategies}
    \vspace{-\intextsep}
\end{figure*}

\section{Methods}

\subsection{Task Definition}

Instruction tuning datasets today are often consisted of carefully curated prompt/response pairs across numerous domain and tasks. In this paper we predominantly experiment with a 90K subset of the FLAN V2 data~\citep{longpre2023flan} and the entire Dolly dataset~\citep{conover2023free} (with responses initially masked to simulate the label-efficient learning scenario).

The FLAN V2 dataset provides a broad and diverse collection of examples across 1,691 tasks, including classification (e.g., amazon polarity user satisfied), summarization (e.g., news editorial summary), translation (e.g., translate/french-english), question answering (e.g., trivia qa), and many others. Similarly, the Dolly Dataset consists of multiple instruction-following tasks across 8 categories, including question answering, classification, brainstorming, information extraction, summarization, and creative writing.
In our experiments, we define tasks as the original subtasks or categories annotated in the dataset, and distribute the annotation budget across these predefined subtasks.

In modern LLM development, task and domain categorizations are widely available. When building SFT datasets, practitioners need to curate specialized prompts for each domain or application need. For example, both enterprise and academic efforts now emphasize domain-aligned SFT to improve performance in specialized settings such as legal reasoning, biomedical QA, customer support, and other specific use cases.

In this paper, we utilize these task categorizations to demonstrate that task-diverse annotation--where specific budgets are allocated to each individual task category--is as powerful as, if not more powerful than, the diversity methods mentioned previously. With a reweighting scheme based on model confidences, our method yields comparable or superior results to previous approaches. Additionally, our approach validates the current practice of curating task-specific subsets of SFT data in LLM development.

As illustrated in Figure~\ref{fig:allocation_strategies}, our method explicitly incorporates task information to guide selection, ensuring that all tasks are represented in the annotated subset. Previous diversity-based selection methods, which rely on embedding space coverage, often overlook task-level imbalances—leading to underrepresentation of tasks with fewer examples or lower model confidence. In contrast, the Weighted Task Diversity strategy adapts allocation based on model confidence, assigning more examples to lower-confidence tasks, such as Task 5.

\subsection{Task Diversity}

As a novel baseline method that leverages task-diversity, we optimize for task-level diversity in the selected subset under a fixed annotation budget. Our objective is simple: allocate roughly equal number of examples to each task subject to the availability of prompts available in each task. We propose a two-step sampling algorithm that first allocates the budget across tasks and then samples examples from each task. This approach is denoted as \textbf{Task Diversity} in our results.

Formally, under a fixed annotation budget \( B \), we formulate the allocation problem as minimizing the maximum number of examples per task while ensuring complete budget utilization and task-specific availability constraints. Formally, let there be \( T \) tasks with their corresponding prompt partitions $X_1, ..., X_T \subset X$. Here, $X_t$ represents the prompts belonging to task $t$. We also let variables \( \boldsymbol{\alpha} = [\alpha_1, \alpha_2, \dots, \alpha_T] \in \mathbb{R}^T\) denote the variable of allocated examples per task. The goal is to solve the following min-max problem:
\begin{align*}
&\boldsymbol{\alpha}_{\text{Div}} = \argmin_{\boldsymbol{\alpha}} \left( \max_{t \in \{1, \dots, T\}} \alpha_t \right)\\
& \text{s.t.} \; \sum_{t=1}^{T} \alpha_t = B \quad \text{and} \quad \forall t\in [T], \alpha_t \leq |X_t|.
\end{align*}
Here, $\boldsymbol{\alpha}_{\text{Div}} \in [0, 1]^T$ denotes the budget allocations. The optimal solution would saturate the tasks with fewer amount of examples, while allocating equal amount of budget to the rest of the tasks. As elements in $\boldsymbol{\alpha}_{\text{Div}}$ may not be integers, we adopt a round robin algorithm for selection as will be described in Section~\ref{ssec:round_robin}.

\subsection{Weighted Task Diversity}
In the second method, our goal is to prioritize sampling from tasks where the model exhibits lower confidence, while still maintaining task-level coverage under a fixed annotation budget. We propose a two-step sampling algorithm that first allocates the budget across tasks based on their average model confidence and then samples examples from each task. This approach is denoted as \textbf{Weighted Task Diversity } in our results.

Under a fixed annotation budget \( B \), we first compute the average confidence score \( \text{conf}_t \) for each task \( t \in \{1, \ldots, T\} \).
We define the average task-level confidence  \( \text{conf}_t \) as the mean model confidence over all unlabeled examples within task \( t \), where confidence for an individual example is computed as the product of token-level probabilities in the generated sequence \citep{settles2009active,bhatt2024experimental}.
Formally, for a given prompt \( x \), let the auto-regressive base model $g$ generate a response \( y = (y_1, y_2, \dots, y_m) \), where \( y_j \) is the \( j \)-th token in the sequence. The model's confidence for this example is defined as:
$\text{conf}(x) = \prod_{j=1}^{m} g(y_j \mid y_{<j}, x)$.
Then, the task-level average confidence can be computed as an average over all prompts:
$\text{conf}_t = \frac{1}{|X_t|} \sum_{x \in X_t} \text{conf}(x)$ for task $t$.

During our selection, each task is initially allocated a small base budget (e.g., 5 examples) to ensure minimum task coverage. The remaining budget is then distributed across tasks with more than five available examples in proportion to the inverse of their confidence scores, thus favoring tasks where the model is more uncertain. Recall that \(X_1, X_2, \dots, X_T \) denote the set of prompt corresponding to each task, and \( \boldsymbol{\alpha} = [\alpha_1, \alpha_2, \ldots, \alpha_T] \) denotes the final allocated samples per task. The allocation rule can be formally written as solving for $C \in \mathbb{R}^+$, 
\begin{align*}
    &\alpha_t = \left[C \cdot \frac{1}{\text{conf}_t}\right]_5^{|X_t|} \quad \text{and}\quad \sum_{t=1}^T \alpha_t = k.
\end{align*}
Where $C$ is a normalization constant used to proportionally distribute the remaining annotation budget among eligible tasks.
In addition, $\big[z\big]_{a}^b := \min(\max(z, a), b)$ denotes the clamping operation, ensuring the number of selected examples per task remains within valid bounds. We use $\boldsymbol{\alpha}_{\text{Weighted}}$ to denote the allocation. Overall, if all tasks have sufficient number of examples and the budget is sufficiently large, the number of examples allocated to each task is exactly proportional to its inverse confidence score.

\begin{algorithm}[t]
\caption{Round-Robin Sampling}
\label{alg:roundrobin}
\begin{algorithmic}[1]
\STATE\textbf{Input: } Budget $k$ and allocation $\boldsymbol{\alpha}$.

\STATE\textbf{Input: } Pool of prompts $X = X_1, ..., X_T$ partitioned based on tasks $1, .., T$ and sorted based on allocation budget so that 
${\alpha}_1 \leq {\alpha}_2 \leq \cdots \leq {\alpha}_T$. 


\STATE \textbf{Initialize: } Selected examples: $S \leftarrow \emptyset$; counters for each task $t$: $c_t \leftarrow 0, \forall t\in[T]$.

\WHILE{True}
    \FOR{t = 1,..., T}
        \IF{$c_t < \lceil \alpha_t \rceil$ and $c_t < |X_t|$}
            \STATE Randomly select example $x$ from $X_t$ that is not in $S$ yet.
            \STATE Update selected set: $S \leftarrow S \cup \{x\}$ and $c_t \leftarrow c_t + 1$
            \STATE \textbf{if} $|S| = B$ \textbf{then} \textbf{break}
        \ENDIF
    \ENDFOR
\ENDWHILE

\RETURN Selection set $S$.
\end{algorithmic}
\end{algorithm}

\subsection{Round Robin}\label{ssec:round_robin}
As we noted before, both $\boldsymbol{\alpha}_{\text{Div}}$ and $\boldsymbol{\alpha}_{\text{Weighted}}$ may yield non-integer allocations. In practice, we round up the values $\lceil\boldsymbol{\alpha}_{\text{Div}}\rceil$ and $\lceil\boldsymbol{\alpha}_{\text{Weighted}}\rceil$, where the rounding operation is computed elementwise. To ensure we only select $k$ examples in total, as shown in Algorithm~\ref{alg:roundrobin}, we adopt a round-robin allocation scheme that iteratively distributes the budget across tasks. We initialize all tasks as eligible for allocation and sequentially assign one example to each task in turn. This process is repeated in a loop, until either the task-specific budget is exhausted or all prompts in a task have all been selected. This procedure continues until the entire budget is fully allocated. During this process, we ensure the smaller budget tasks are prioritized in receiving examples. Within each task, we simply choose prompts uniformly without replacement.

\begin{table*}[ht]
\centering
\small
\setlength{\tabcolsep}{3pt} 
\begin{tabular}{@{}l|cccc|cccc@{}}
\toprule
& \multicolumn{4}{c|}{\textbf{MMLU Score}} & \multicolumn{4}{c}{\textbf{BBH Score}} \\
\textbf{Strategy} & \textbf{k = 20K} & \textbf{k = 30K} & \textbf{k = 45K} & \textbf{k = 90K} & \textbf{k = 20K} & \textbf{k = 30K} & \textbf{k = 45K} & \textbf{k = 90K} \\
\midrule
Random & $45.67_{\pm0.04}$ & $46.28_{\pm0.48}$ & $47.85_{\pm0.10}$ & $48.94_{\pm0.17}$ & $38.42_{\pm0.20}$ & $39.99_{\pm0.40}$ & $38.14_{\pm1.17}$ & $40.02_{\pm0.24}$ \\
Mean Entropy & $43.79_{\pm0.33}$ & $45.49_{\pm0.25}$ & $46.82_{\pm0.27}$ & $48.94_{\pm0.17}$ & $38.45_{\pm0.26}$ & $38.27_{\pm0.65}$ & $40.14_{\pm0.27}$ & $40.02_{\pm0.24}$ \\
Confidence & $45.24_{\pm0.24}$ & $45.70_{\pm0.25}$ & $47.20_{\pm0.24}$ & $48.94_{\pm0.17}$ & $39.50_{\pm0.69}$ & $38.68_{\pm0.82}$ & $39.55_{\pm0.20}$ & $40.02_{\pm0.24}$ \\
Mean Margin & $44.99_{\pm0.03}$ & $46.18_{\pm0.08}$ & $47.22_{\pm0.18}$ & $48.94_{\pm0.17}$ & $39.40_{\pm0.06}$ & $39.65_{\pm0.45}$ & $39.86_{\pm0.47}$ & $40.02_{\pm0.24}$ \\
Min Margin & $46.09_{\pm0.11}$ & $46.51_{\pm0.06}$ & $47.65_{\pm0.24}$ & $48.94_{\pm0.17}$ & $\underline{40.44}_{\pm0.48}$ & $40.42_{\pm0.42}$ & $38.68_{\pm0.48}$ & $40.02_{\pm0.24}$ \\
k-Center & $45.83_{\pm0.16}$ & $46.47_{\pm0.14}$ & $47.74_{\pm0.14}$ & $48.94_{\pm0.17}$ & $38.65_{\pm0.29}$ & $38.94_{\pm0.53}$ & $38.32_{\pm0.77}$ & $40.02_{\pm0.24}$ \\
FL($\gamma$=0.1) & $44.01_{\pm0.23}$ & $45.31_{\pm0.92}$ & $46.47_{\pm0.17}$ & $48.94_{\pm0.17}$ & $37.40_{\pm0.13}$ & $38.17_{\pm0.68}$ & $40.09_{\pm0.83}$ & $40.02_{\pm0.24}$ \\
FL($\gamma$=0.002) & $44.73_{\pm0.21}$ & $46.92_{\pm0.43}$ & $47.71_{\pm0.20}$ & $48.94_{\pm0.17}$ & $38.91_{\pm0.40}$ & $40.70_{\pm0.46}$ & $\mathbf{41.68}_{\pm0.34}$ & $40.02_{\pm0.24}$ \\
FL(cosine) & $44.49_{\pm0.09}$ & $45.18_{\pm0.27}$ & $46.51_{\pm0.24}$ & $48.94_{\pm0.17}$ & $39.53_{\pm0.35}$ & $39.27_{\pm1.02}$ & $40.40_{\pm0.22}$ & $40.02_{\pm0.24}$ \\
ActiveIT & $\underline{47.50}_{\pm0.15}$ & $\underline{47.56}_{\pm0.25}$ & $47.66_{\pm0.24}$ & $48.94_{\pm0.17}$ & $39.52_{\pm0.11}$ & $39.88_{\pm0.08}$ & $40.78_{\pm0.59}$ & $40.02_{\pm0.24}$ \\
DPP & $45.58_{\pm0.15}$ & $46.87_{\pm0.10}$ & $47.84_{\pm0.26}$ & $48.94_{\pm0.17}$ & $\mathbf{40.68}_{\pm0.50}$ & $\underline{40.99}_{\pm0.47}$ & $40.07_{\pm0.41}$ & $40.02_{\pm0.24}$ \\

Task Diversity & $45.83_{\pm0.57}$ & $46.25_{\pm0.12}$ & $\mathbf{48.63}_{\pm0.18}$ & $48.94_{\pm0.17}$ & $39.02_{\pm0.64}$ & $39.63_{\pm0.36}$ & $\underline{41.10}_{\pm0.53}$ & $40.02_{\pm0.24}$ \\

\shortstack[l]{Weighted Task\\Diversity} & $\mathbf{47.88}_{\pm0.19}$ & $\mathbf{48.34}_{\pm0.16}$ & $\underline{48.46}_{\pm0.15}$ & $48.94_{\pm0.17}$ & $39.96_{\pm0.52}$ & $\mathbf{41.04}_{\pm0.33}$ & $40.86_{\pm0.15}$ & $40.02_{\pm0.24}$ \\
\bottomrule
\end{tabular}
\caption{Comparison of model performance on MMLU (left) and BBH (right) benchmarks using different data selection strategies and annotation budgets. Each result is averaged over 3 random seeds where the randomness mainly comes from the training. The confidence intervals are based on standard error. The best results for each k are in \textbf{bold}, and the second-best results are \underline{underlined}.}
\label{tab:flan-results}
\end{table*}
\section{Experiments}
\subsection{Experiment Setup}


\noindent\textbf{Dataset. }
We utilize a curated 90K subset of the FLAN V2 dataset \citep{longpre2023flan}, as processed by~\citet{wang2023far}. FLAN V2 is an instruction finetuning dataset that integrates data from multiple sources, including FLAN 2021, P3++, Super-Natural Instructions, and a range of additional datasets focused on reasoning, dialogue, and program synthesis. 
In addition, we use the Databricks Dolly 15K dataset~\citep{conover2023free}, which comprises 15,000 human-generated instruction-following examples. The dataset was crowd-sourced from thousands of Databricks employees, who authored prompt-response pairs across eight instruction categories

\noindent\textbf{Models and Training Procedure.} We conduct our experiments using the 7B parameter version of the LLaMA-2 language model~\citep{touvron2023llama}, considering various annotation budgets. Before finetuning, we select a subset of prompts for annotation using the baseline algorithms described below and our \textbf{Task Diversity}, \textbf{Weighted Task Diversity} algorithms. These strategies are computed based on the original, pre-trained model only. After selection, we finetune the model on the annotated prompt-response pairs using Low-rank Adaptation~\citep{hu2021lora}. We finetune each model for 3 epochs using the Adam optimizer with a learning rate of $ 10^{-4}$.

\noindent\textbf{Baseline Algorithms.} We evaluate a diverse set of baseline algorithms for prompt selection mostly adopted from \citet{kung2023active}, \citet{bhatt2024experimental} and \citet{wang2024diversity}. \textbf{Random} represents the random sampling baseline.
We also include \textbf{Mean Entropy}, which measures the average tokenwise negative entropy of the softmax probability scores in the generated sequence; \textbf{Least Confidence}, which selects prompts with the lowest overall sequence probability, computed as the product of the token probabilities; \textbf{Mean Margin}, which captures uncertainty as the average difference between the highest and second-highest token probabilities at each position in the sequence; and \textbf{Min Margin}, a more targeted variant that considers the smallest such margin across all tokens, rather than the average.
Furthermore, the task-wise allocation method \textbf{ActiveIT}~\citep{kung2023active} is also included. In this method, algorithm selects tasks where the model exhibits the lowest average confidence and annotates all examples from these tasks.

Lastly, we also include diversity-based selection methods such as \textbf{K-Center}, \textbf{Facility Location} and \textbf{DPP} (detailed in Section~\ref{ssec:div_review}).
\begin{figure*}[t]
    \begin{minipage}{0.64\textwidth}
        \centering
        \includegraphics[width=\textwidth]{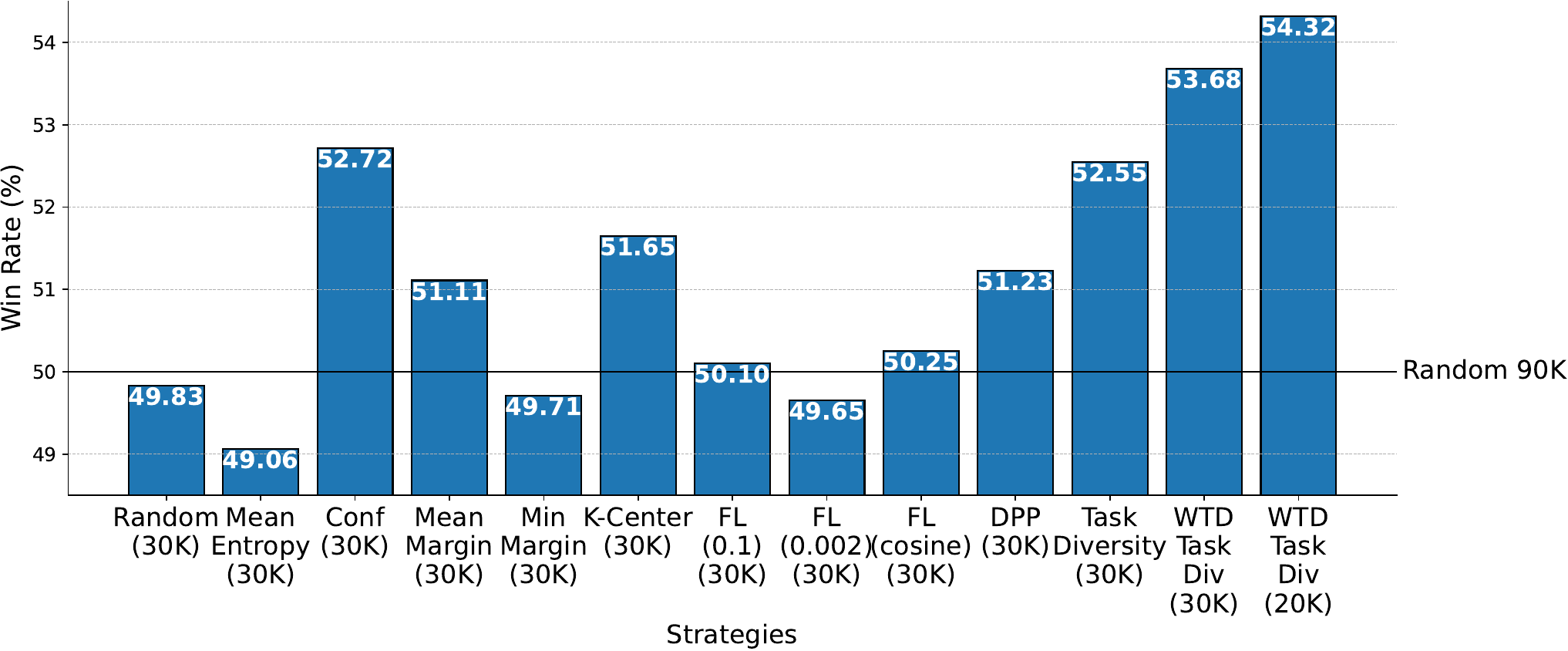}
        \caption{Evaluation of 30K prompt selection strategies on the FLAN V2 dataset using AlpacaEval. The win rate represents the proportion of times each model's response was preferred by GPT-4 Turbo over the 90K random baseline. ''WTD Task Div" here indicates Weighted Task Diversity and ''Conf" indicates Confidence}
        \label{fig:alpacaeval-flan}
    \end{minipage}
    \hfill
    \begin{minipage}{0.35\textwidth}
        \centering
        \setlength{\tabcolsep}{2pt} 
        \scalebox{0.7}{
        \begin{tabular}{lccc}
        \toprule
        \textbf{Strategy} & \textbf{k = 3K} & \textbf{k = 6K} & \textbf{k = 13.5K} \\
        \midrule
        Random & $33.33_{\pm0.49}$ & $34.65_{\pm0.84}$ & $35.33_{\pm1.03}$ \\
        Mean Entropy & $\underline{34.35}_{\pm0.31}$ & $33.74_{\pm0.76}$ & $35.33_{\pm1.03}$ \\
        Confidence & $34.29_{\pm1.71}$ & $37.51_{\pm0.60}$ & $35.33_{\pm1.03}$ \\
        Mean Margin & $32.61_{\pm1.96}$ & $33.01_{\pm0.74}$ & $35.33_{\pm1.03}$ \\
        Min Margin & $29.43_{\pm0.35}$ & $33.85_{\pm1.22}$ & $35.33_{\pm1.03}$ \\
        k-Center & $33.61_{\pm0.85}$ & $32.90_{\pm1.17}$ & $35.33_{\pm1.03}$ \\
        FL($\gamma$ = 0.1) & $33.19_{\pm2.38}$ & $33.11_{\pm0.15}$ & $35.33_{\pm1.03}$ \\
        FL($\gamma$ = 0.002) & $31.70_{\pm1.73}$ & $33.22_{\pm1.76}$ & $35.33_{\pm1.03}$ \\
        FL(cosine) & $34.03_{\pm1.97}$ & $\underline{35.52}_{\pm0.66}$ & $35.33_{\pm1.03}$ \\
        DPP & $31.02_{\pm1.49}$ & $32.70_{\pm1.29}$ & $35.33_{\pm1.03}$ \\
        Task diversity & $33.62_{\pm0.34}$ & $32.65_{\pm0.42}$ & $35.33_{\pm1.03}$ \\
        \shortstack[l]{Weighted Task\\Diversity} & $\mathbf{39.74}_{\pm0.54}$ & $\mathbf{38.93}_{\pm0.34}$ & $35.33_{\pm1.03}$ \\
        \bottomrule
        \end{tabular}
        }
        \captionof{table}{MMLU evaluation of models trained on subsets selected from a pool of 13.5K examples from the Dolly dataset. Each result is averaged over 3 trials.}
        \label{dolly-mmlu}
    \end{minipage}
\end{figure*}

\noindent\textbf{Evaluation Metrics.}
We evaluate our finetuned model’s zero-shot capabilities using three benchmarks: MMLU (Massive Multitask Language Understanding, \citet{hendrycks2020measuring}), BBH (Big-Bench Hard, \citet{suzgun2022challenging}), and AlpacaEval~\citep{alpaca_eval}. Following FLAN V2’s methodology, MMLU tests factual knowledge and reasoning across 57 subjects via multiple-choice questions spanning elementary to professional levels, while BBH evaluates general reasoning capabilites of the model through 23 generation-based.
AlpacaEval is an evaluation framework designed to automatically measure the quality of responses generated by large language models in instruction-following tasks. It compares outputs from two models to determine which response better follows the given instruction. In our study, we used AlpacaEval 2.0 with the GPT-4 Turbo model as the annotator, which ranks the outputs from our finetuned model (trained on a subset of data) against those from a reference model (trained on the full dataset). For our results, we report the length controlled win rates, accounting for potential biases of long answer preferences by GPT-4 Turbo.

\subsection{Evaluation on MMLU and BBH}
In Table~\ref{tab:flan-results}, when experimenting with the FLAN V2 dataset, we observe that Weighted Task Diversity substantially outperform traditional sampling methods on the MMLU benchmark under different annotation budgets. At $45K$ budget, models trained based on Weighted Task Diversity selection achieves similar performance to the model trained on all $90K$ examples in the dataset, effectively saving $50\%$ annotation budget when compared to random sampling. Overall, this demonstrates that task-aware sampling, particularly when informed by model uncertainty, can achieve near-optimal performance even under low budgets. On the BBH benchmark, however, performance varies more significantly across strategies and budgets, with no single method consistently dominating. However, at 30K budget, our strategy achieves the best overall performance with a score surpassing the accuracy when labeling all $90K$ examples.
 
In Table~\ref{dolly-mmlu}, we also include the performance of Weighted Task Diversity on the Dolly dataset. Our method achieves the highest MMLU scores at both 3K and 6K budgets, reaching $39.74\%$ and $38.93\%$ accuracies, respectively. Notably, as we are selecting a more informative set and balanced set of examples, the MMLU score even outperforms training on all examples by more than $4\%$ and saving $80\%$ in annotation budget. We also include the BBH result for the Dolly dataset in Table~\ref{bbh-dolly} of Appendix~\ref{sec:appendix}. However, even with random sampling, we see the model performance consistently dropping as more examples are being labeled and trained on. This potentially suggests the ineffectiveness of the Dolly dataset in improving the tasks covered by BBH, suggesting us to downweight the significance of this benchmark.

\subsection{Evaluation by GPT-4}
To further investigate the effectiveness of various data selection strategies under constrained annotation budgets, we use the finetuned models on 30K prompts for all strategies and compare them against a baseline model trained on 90K randomly selected prompts. Evaluation is conducted using GPT-4 Turbo as the judge following the AlpacaEval framework~\citep{alpaca_eval}, having over 805 prompts. As shown in Figure~\ref{fig:alpacaeval-flan}, several strategies surpass the Random (90K) baseline, despite using significantly fewer examples.

Among all strategies, Weighted Task Diversity 20K and 30K budget achieve the highest win rates of 54.32\% and 53.68\% for FLAN Datset. Even on the Dolly Dataset as shown in Figure~\ref{fig:alpacaeval-dolly}, Weighted Task Diversity with 3K and 6K budgets achieves the highest win rates of 52.28\% and 51.17\%, respectively, outperforming all other strategies.

Overall, these evidences suggest that the simple method of Weighted Task Diversity is comparable if not better than the existing baseline methods.

\subsection{Qualitative Analysis}
\begin{figure}[ht]
    \centering
    \includegraphics[width=0.95\linewidth]{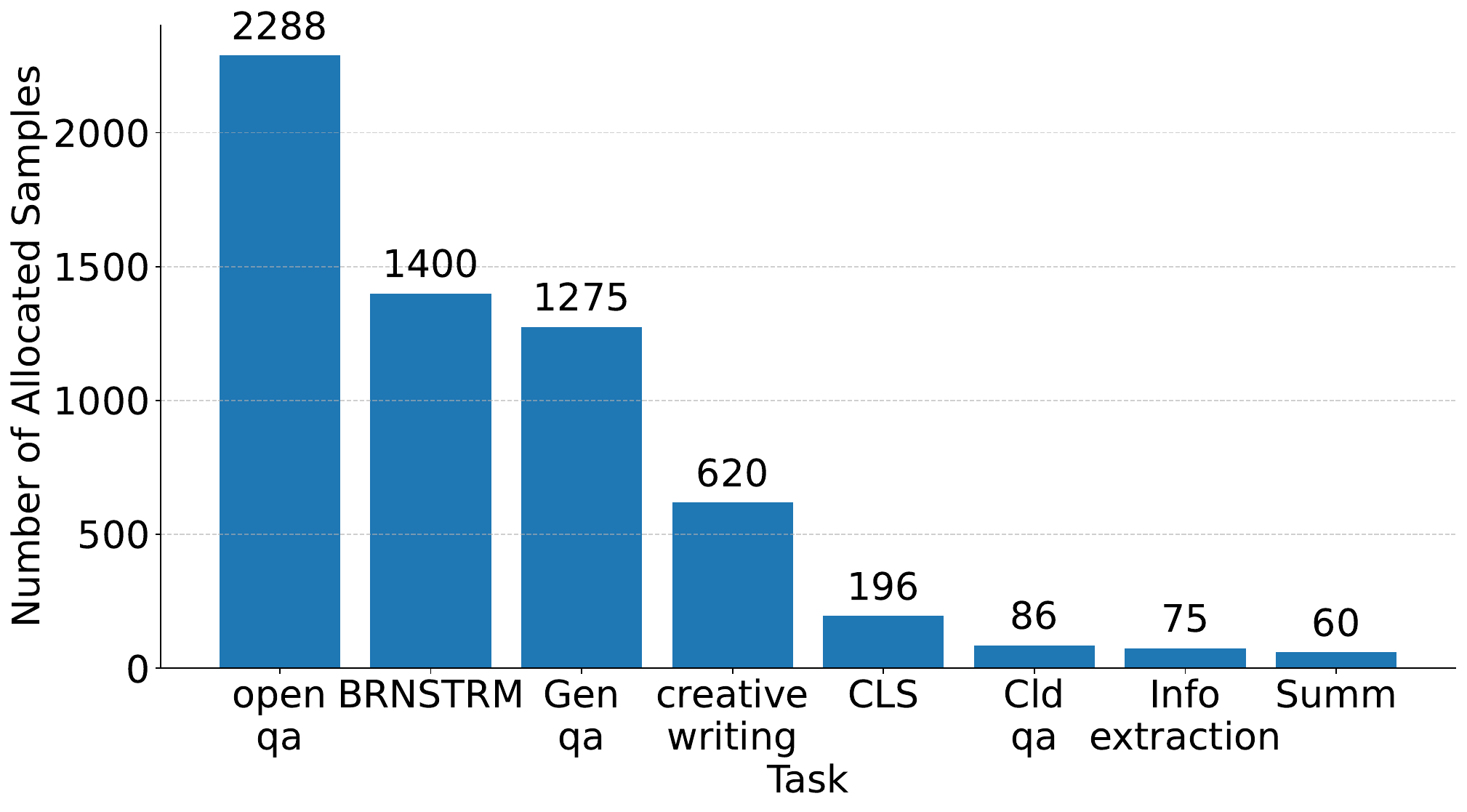}
    \caption{Number of allocated samples per task in the Dolly dataset under the Weighted Task Diversity strategy. ``BRNSTRM" indicates Brainstorming Tasks, ``CLS" indicates Classification Tasks, ``Cld qa" indicates Closed QA and ``Summ" indicates Summarization.}
    \label{fig:allocation-dolly-new}
\end{figure}
In Figure~\ref{fig:allocation-dolly-new}, we plot the task-wise allocation budget of the \textbf{Weighted Task Diversity} strategy for the Dolly dataset.
As we can observe, a majority of samples are allocated to open QA, brainstorming, and general QA tasks. This distribution aligns with intuitive expectations: tasks like open QA typically present greater challenges for pre-trained language models due to their open-ended nature and broad variability in possible responses. 
The relatively lower model confidence on QA-like tasks leads to greater sampling from these categories, consistent with the inverse-confidence weighting mechanism. In contrast, tasks such as classification, summarization, and information extraction receive far fewer samples, likely because they are structurally simpler and the model exhibits higher confidence on them. However, unlike the method proposed by \citet{kung2023active}, these tasks still receive a few annotations, ensuring our dataset has a broad coverage over them. See Appendix~\ref{apx:qualitative} for more information on the FLAN dataset.

\section{Conclusion}

In this paper, we present a simple yet effective approach to label-efficient supervised finetuning by leveraging task-level diversity and model confidence. We introduce two  algorithms--Task Diversity and Weighted Task Diversity--that allocate annotation budgets across tasks based on either uniform distribution or inverse model confidence scores. Our methods consistently outperform or match more complex diversity- and uncertainty-based baselines across MMLU and GPT-4-based AlpacaEval benchmarks, all while using significantly fewer labeled examples. These results highlight the value of leveraging task structure and model uncertainty for cost-effective and scalable instruction tuning of large language models.


\section*{Limitations}
Our approach demonstrates strong performance in label-efficient learning scenarios, though we acknowledge several areas for future exploration. While task information is widely available in modern instruction-tuning datasets as we've shown, there may be specific domains where such categorizations are less defined. In these cases, automated methods for identifying task definitions and categorizations could be beneficial.

The confidence assessment by the base model may sometimes reflect its pre-training exposure rather than inherent task difficulty. However, our experimental results suggest this concern is largely mitigated by our round-robin sampling strategy that ensures minimum coverage across all tasks.

Our experiments primarily focus on the LLaMA-2 7B architecture, and extending these findings to other model architectures represents a promising direction for future work. Additionally, while our method is significantly more computationally efficient than existing diversity-based approaches, calculating confidence scores does require inference on the unlabeled dataset.

\section*{Acknowledgments}
This work has been supported in part by NSF Award 2112471.


\newpage
\bibliography{custom}

\clearpage

\appendix
\onecolumn

\section{Additional Results}
\label{sec:appendix}

\begin{figure*}[htbp]
    \centering
    \includegraphics[width=\textwidth]{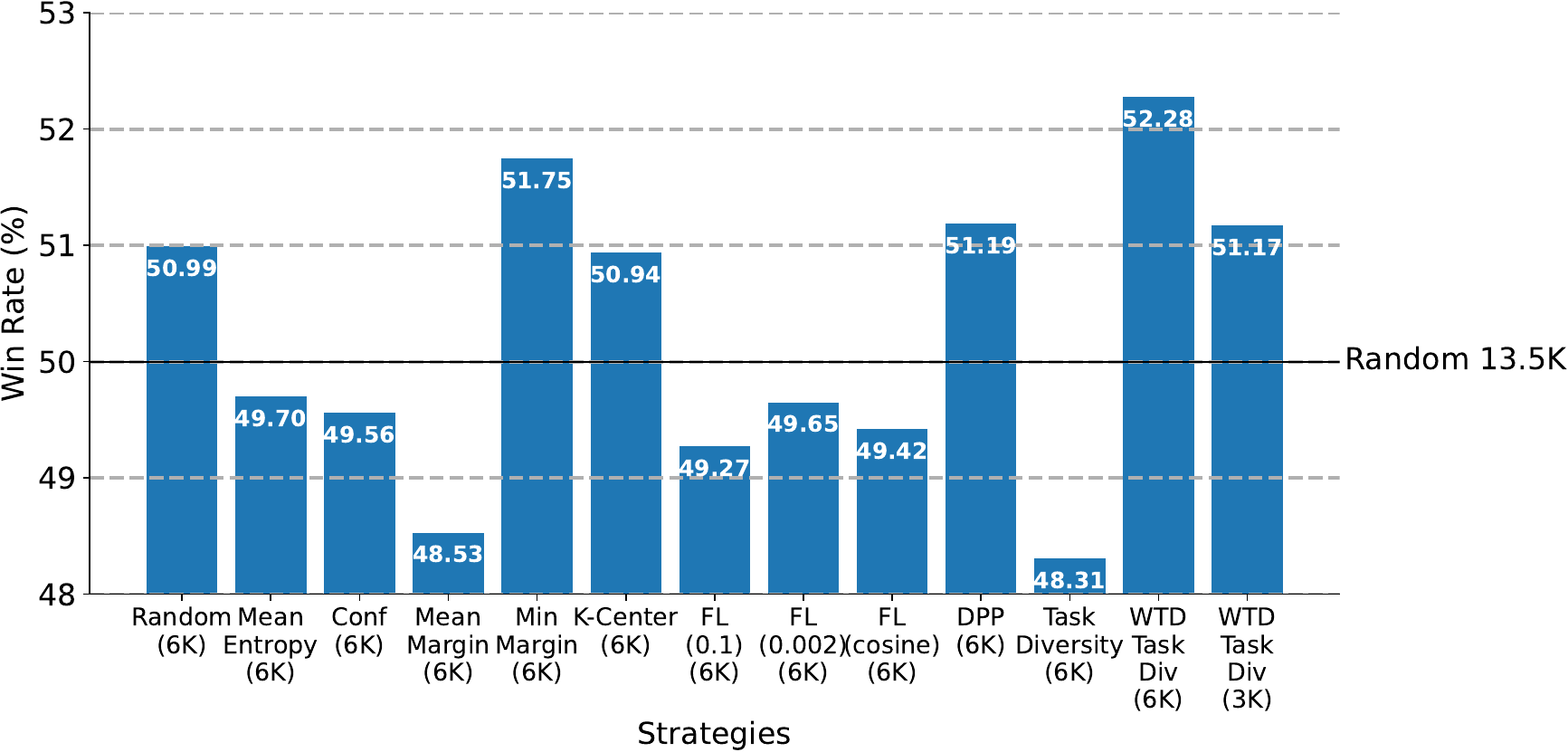}
    \caption{Evaluation of 6K prompt selection strategies on the Dolly dataset using AlpacaEval. The win rate represents the proportion of times each model’s response was preferred by GPT-4 Turbo over the 13.5K random baseline. ''WTD Task Div" here indicates Weighted Task Diversity and ''Conf" indicates Confidence}
    \label{fig:alpacaeval-dolly}
\end{figure*}

\begin{table*}[htbp]
\centering
\setlength{\tabcolsep}{3pt} 
\begin{tabular}{lccc}
\toprule
\textbf{Strategy} & \textbf{k = 3K} & \textbf{k = 6K} & \textbf{k = 13.5K} \\
\midrule
Random & $36.43_{\pm0.11}$ & $35.60_{\pm0.72}$ & $34.02_{\pm0.84}$ \\
Mean Entropy & $35.48_{\pm0.79}$ & $34.07_{\pm0.36}$ & $34.02_{\pm0.84}$ \\
Confidence & $36.63_{\pm0.40}$ & $35.68_{\pm0.55}$ & $34.02_{\pm0.84}$ \\
Mean Margin & $35.79_{\pm0.71}$ & $35.99_{\pm0.37}$ & $34.02_{\pm0.84}$ \\
Min Margin & $36.32_{\pm0.28}$ & $33.89_{\pm1.02}$ & $34.02_{\pm0.84}$ \\
k-Center & $\mathbf{38.99}_{\pm0.45}$ & $35.43_{\pm0.65}$ & $34.02_{\pm0.84}$ \\
FL($\gamma$ = 0.1) & $35.76_{\pm0.28}$ & $\mathbf{37.12}_{\pm0.24}$ & $34.02_{\pm0.84}$ \\
FL($\gamma$ = 0.002) & $35.84_{\pm0.20}$ & $35.89_{\pm0.62}$ & $34.02_{\pm0.84}$ \\
FL(cosine) & $34.84_{\pm0.90}$ & $35.48_{\pm0.02}$ & $34.02_{\pm0.84}$ \\
DPP & $\underline{37.27}_{\pm0.27}$ & $\underline{36.38}_{\pm0.30}$ & $34.02_{\pm0.84}$ \\
Task diversity & $34.96_{\pm0.54}$ & $35.60_{\pm0.38}$ & $34.02_{\pm0.84}$ \\
Weighted task diversity & $37.07_{\pm0.30}$ & $35.63_{\pm0.45}$ & $34.02_{\pm0.84}$ \\
\bottomrule
\end{tabular}
\caption{Big Bench Hard (BBH) evaluation of models trained on subsets selected by strategies from a pool of 13.5K under different annotation budgets on the Dolly dataset. Each result is averaged over 3 random seeds.}
\label{bbh-dolly}
\end{table*}



\section{Qualitative Analysis} 
\label{apx:qualitative}
\begin{figure}[t]
    \centering
    \includegraphics[width=0.8\linewidth]{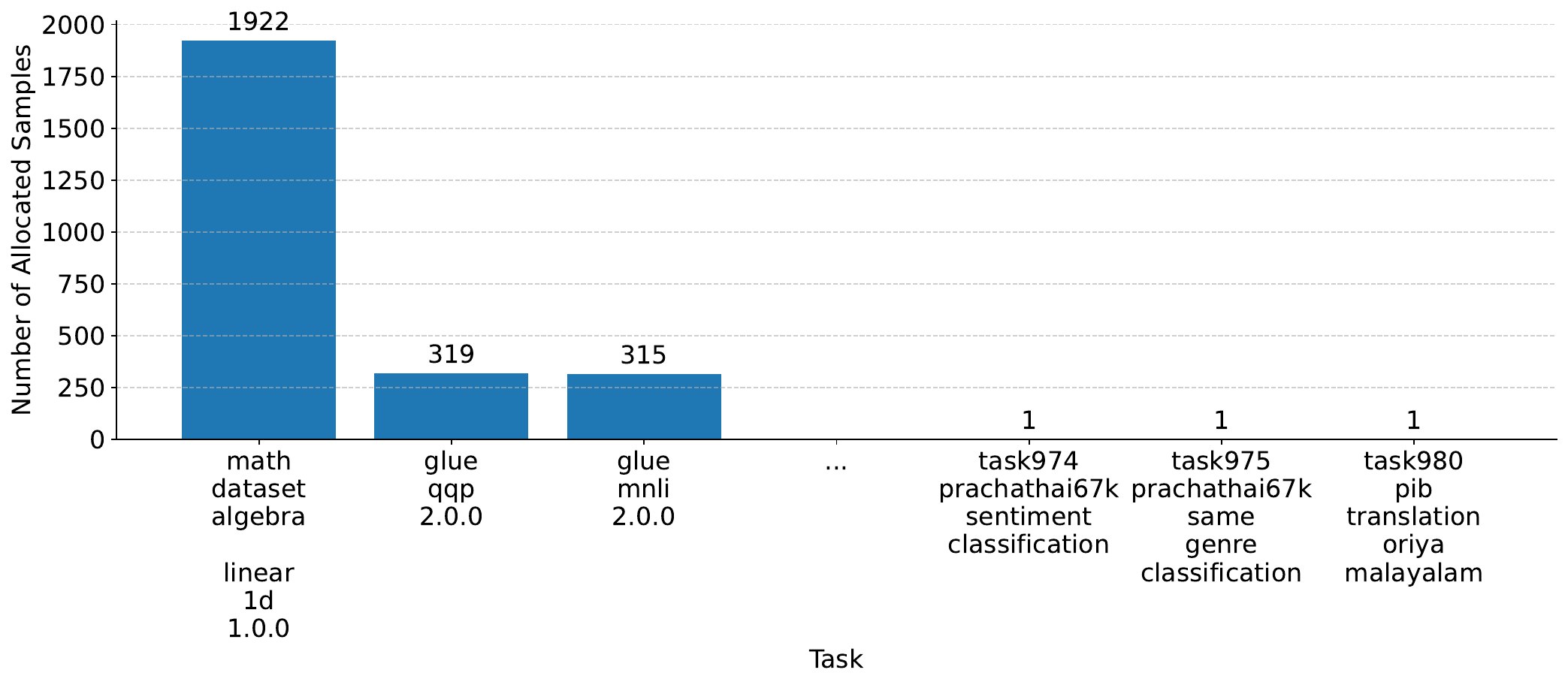}
    \caption{Task-wise sample allocation in the \textbf{FLAN V2} dataset using the \textbf{Weighted Task Diversity} strategy. Tasks where the model exhibits greater uncertainty receive a larger portion of the annotation budget.}
    \label{fig:allocation-flan}
\end{figure}

In Figure~\ref{fig:allocation-flan}, we visualize the task-wise allocation under the \textbf{Weighted Task Diversity} strategy for the FLAN dataset. A significant portion of the annotation budget is allocated to math-related tasks such as math dataset algebra, where pre-trained language models tend to exhibit high uncertainty. Moderate allocation is also seen for paraphrase detection tasks (e.g., Quora Question Pairs(QQP) and MNLI(Multi-Genre Natural Language Inference), which often require nuanced semantic understanding. In contrast, although we set a base allocation of 5 examples per task, some tasks such as prachathai67k sentiment classification and low-resource translation contain only a single available example and are selected accordingly. This behavior reflects the influence of task size constraints rather than model confidence. Nevertheless, such tasks still receive limited annotations, helping to preserve broad task-level coverage and ensuring that all tasks are represented in the final training set.

\section{Additional Experiment Details}
\section*{Licenses for Models and Datasets}
\label{sec:licenses}

The following licenses apply to the models and datasets used in this work:

\begin{itemize}
    \item \textbf{Llama-2 7B}: Released by Meta under a custom non-commercial license.\footnote{\url{https://ai.meta.com/resources/models-and-libraries/llama-downloads/}}
    \item \textbf{FLAN V2 Dataset}: Released by Google under the Apache License 2.0.\footnote{\url{https://github.com/google-research/FLAN/tree/main/flan/v2}}
    \item \textbf{Dolly Dataset}: Released by Databricks under the Creative Commons Attribution-ShareAlike 3.0 (CC BY-SA 3.0) license.\footnote{\url{https://github.com/databrickslabs/dolly}}
\end{itemize}

\section*{Computational Complexity}

We primarily used L40 GPUs. We precompute the embeddings which takes 12 hours. For FLAN Dataset using L40, each trial takes roughly 35 GPU hours including the evaluation for 90K budget, 23 GPU hours for 45K, 18 hours for 30K and 15 GPU hours for 20K respectively. For Dolly Dataset using L40, each trial takes roughly 10 GPU hours including the evaluation for 13.5K budget, 9 GPU hours for 6K, 8 GPU hours for 3K respectively.

\end{document}